\documentclass{article}
\usepackage{spconf,amsmath,graphicx,booktabs}
\usepackage{array,multirow, hyperref}
\usepackage{lipsum}

\title{A Convolutional Autoencoder Approach to Learn Volumetric Shape Representations for Brain Structures}
\name{Evan M. Yu and Mert R. Sabuncu}
\address{Meinig School of Biomedical Engineering and School of Electrical and Computer Engineering \\ Cornell University, Ithaca, NY 14853}

\begin{document}
%
\maketitle
\begin{abstract}
We propose a novel machine learning strategy for studying neuroanatomical shape variation. 
Our model works with volumetric binary segmentation images, and requires no pre-processing such as the extraction of surface points or a mesh.
The learned shape descriptor is invariant to affine transformations, including shifts, rotations and scaling. 
Thanks to the adopted autoencoder framework, inter-subject differences are automatically enhanced in the learned representation, while intra-subject variances are minimized. 
Our experimental results on a shape retrieval task showed that the proposed representation outperforms a state-of-the-art benchmark for brain structures extracted from MRI scans. 

\end{abstract}
\begin{keywords}
shape analysis, deep learning, spatial transformers, autoencoder. \let\thefootnote\relax\footnotetext{This work has been submitted to the IEEE for possible publication. Copyright may be transferred without notice, after which this version may no longer be accessible}
\end{keywords}
\section{Introduction}
\label{sec:intro}

Over the last two decades, neuroimaging has revolutionized our understanding of brain anatomy by allowing us to examine population variation at an unprecedented scale. 
One aspect of brain morphology that has received considerable attention is \textit{shape}.
Shape, in general, refers to the geometric properties of an object (e.g., a brain structure or a region of interest) that are independent of size or volume. 


A broad range of techniques have been developed for the study of shapes of brain structures, which are under significant genetic influence~\cite{ge2016multidimensional} and can yield sensitive biomarkers of disease.
A thorough review is provided by Ng \textit{et al.} \cite{ng2014shape}. 
Following their convention, shape analysis techniques can be broadly grouped into five distinct types. 
One group covers techniques that work on point-based or local features \cite{point1}, whereas another category includes methods that are surface based \cite{shapedna, brainprint}. 
A third category utilizes basis functions such as spherical harmonics \cite{spherical} to represent the geometry, while a fourth group includes skeleton based schemes, such as medial profiles \cite{medial}. 
Finally, there is a category of methods that rely on characterizing deformations \cite{currents}.

Today, most aforementioned techniques would be considered hand-crafted, as they heavily rely on arbitrary modeling choices in order to extract representations.
Recently, deep neural networks have revitalized so-called end-to-end learning approaches that discover optimal representations in a data-driven fashion. 
In this work, we present an unsupervised approach to learn shape representations of different brain regions. 
In our framework, rotation, scaling and translation are normalized via a spatial transformer network \cite{stn} that aligns an input shape to a population template. 
Our shape template is not arbitrary, but also learned during training. 
Finally, the aligned structure is fed to an autoencoder that learns to encode the input into a shape descriptor. 
The network is trained in an end-to-end fashion with binary segmentation volumes as input, and requires no other preprossessing.  
We report results for shape retrieval experiments in the OASIS dataset \cite{oasis}.

The rest of the paper is organized as follows. 
Section 2 discusses machine learning based approaches closely related to our approach. 
Section 3 introduces the proposed unsupervised learning strategy. 
Section 4 presents and discusses the empirical results. 
Finally, Section 5 concludes our paper. 

\section{Machine Learning based Shape Analysis}
\label{sec:related}
There has been a recent surge in the use of machine learning techniques to derive shape features. 
Some basic approaches include projecting the 3D objects onto different views before submitting these to a convolutional neural network (CNN) \cite{projection3, projection2, projection4}. 
In this framework, the user needs to choose an arbitrary set of views, which can be sub-optimal for characterizing a 3D object.

An alternative approach is to represent objects as 3D point clouds, which are then processed using a discriminative neural network such as the PointNet \cite{point1, point2}. 
However, this strategy yields representations that are optimal for a specific task, as in predicting Alzheimer's disease.  

Another set of techniques rely on extracted surface meshes. 
For example, one can compute a heat kernel signature (HKS) of a mesh, which is then fed to an autoencoder~\cite{hks1,hks2}.  
The HKS features are not scale invariant by design. In addition, techniques used in ~\cite{hks1,hks2} are for object classification, but not instance retrieval, which is our focus in this study.
A different mesh-based approach was recently presented by Shakeri \textit{et. al} \cite{shakeri2016deep}, who used a spectral matching method to establish pointwise correspondence across samples and a hybrid auto-encoder and discriminator strategy to learn representations that are optimal for classifying subjects. 
More recently, there has been a growing effort in generalizing deep learning algorithms to non-Euclidean data such as those on mesh graphs or manifolds.
Some of these algorithms have been applied to shape analysis \cite{geo1,geo2}, which are collectively referred to as Geometric Deep Learning.
The main drawback of mesh-based techniques is that the quality of the representation strongly depends on the quality of the surface mesh, which can suffer from topological errors.

Another approach related to our work is the 3D ShapeNet \cite{wu20153d}, which was originally developed to handle 2.5D depth data, and yield shape representations optimized for object class recognition.
To our knowledge, ShapeNet has not been applied to the shape analysis of neuroanatomical structures yet.
Furthermore, this strategy does not have isometry or affine invariance built into the model. 


\begin{figure*}[h]
\begin{minipage}[b]{1.0\linewidth}
  \centering
  \centerline{\includegraphics[width=12cm]{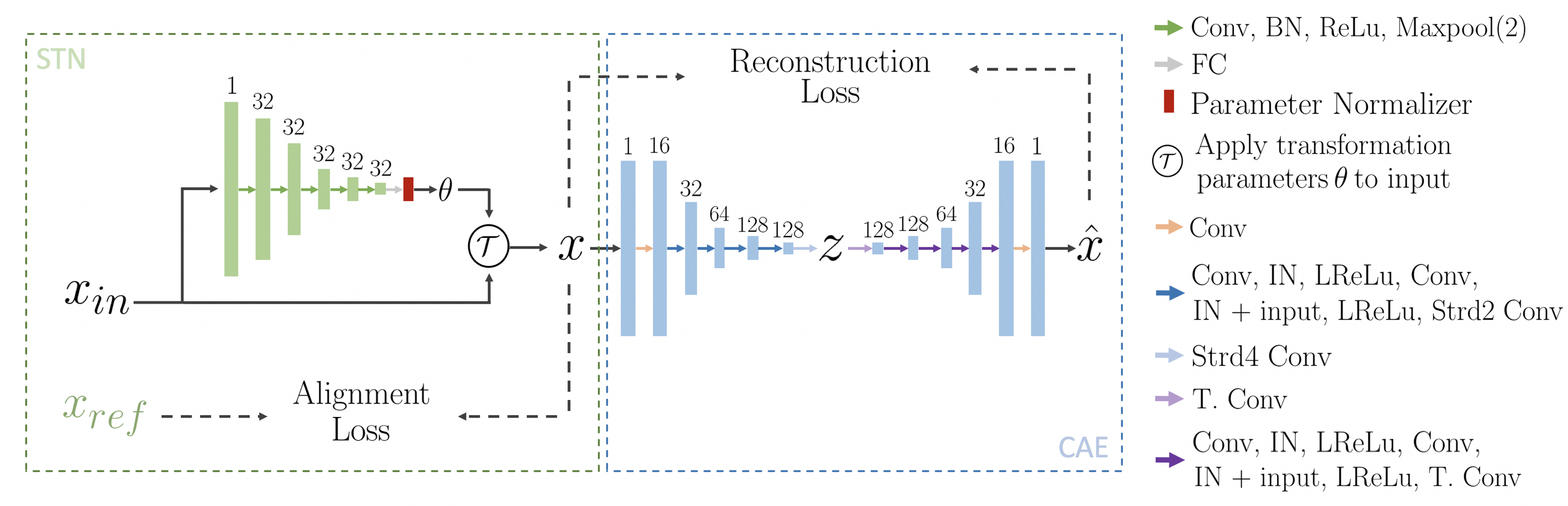}}
\end{minipage}
\caption{Proposed architecture. The network consists of a spatial transformer network (STN) and a convolutional autoendoer (CAE). 
The STN takes input $x_{in}$, a binary segmentation volume, and computes a set of affine transformation parameters $\theta$, which are used to align to the learned reference template $x_{ref}$ using the affine transformation $\mathcal{T}$. 
The template-aligned scan $x$ is passed through a CAE in order to obtain a shape descriptor $z$ from its bottleneck. 
The CAE has several residual blocks, where ``+input'' in the legend indicates a skip connection.
Conv:for a $3\times3$ convolution, IN: Instance normalization, LReLU: Leaky Rectified Linear Unit, T.Conv: Transposed convolution, Strd2 Conv: convolution with stride 2. The number of channels is indicated above each layer.} 
\label{fig:arch}
\end{figure*}

\section{Proposed Method}
\label{sec:method}
Our method obtains a 3D shape descriptor without the need to extract a point cloud or mesh representation of the structure boundary.
The learned shape descriptor is invariant to rotation, translation, and axis-independent scaling. 
Our proposed architecture consist of two components, namely a spatial transformer network (STN) \cite{stn} and a convolutional autoencoder (CAE), as illustrated in Figure \ref{fig:arch}. 
We achieve invariance against affine transformations (excluding shear) via the STN, which aligns the input 3D segmentation $x_{in}$ to a structure-specific reference template $x_{ref}$. 
Note that this template is not pre-set, but learned during training via minimizing the loss function described below.
In the STN component, a convolutional neural network computes 9 transformation parameters (collectively denoted as $\theta$) that make up an affine transformation matrix.
These parameters are three rotation angles, three translations (axis aligned shifts) and three (axis specific) scales. 
Note that we did not include shearing in our transformation model as we considered this to affect the ``shape.''
Next, the output of the STN is converted into a transformation matrix $\mathcal{T}$, which is then applied to the image grid, producing a deformed sampling grid. 
The sampling grid defines where on the input image to sample in order to produce a template-aligned output $x$. 
Finally, the model passes $x$ to an CAE, which goes through a bottleneck representation, namely the shape descriptor $z$, before decoding it into a reconstruction $\hat{x}$. 
The entire model (STN and CAE) is trained end-to-end, minimizing the following loss function:
\begin{equation}
    \mathcal{L}= -\textup{Dice}(x_{in},x_{ref}) - \zeta(t) \textup{Dice}(x,\hat{x}). \label{eq:loss}
\end{equation}
Dice is a commonly used metric that quantifies the similarity between two segmentation maps (e.g., binary volumes).
In our implementation, we treated $x_{in}, x_{ref}$, and $\hat{x}$ as probabilistic segmentations, where each voxel took a value between $0$ and $1$, indicating the probability of the structure of interest.
Hence, we defined the Dice metric as the sum of the voxel-wise product of the two input volumes divided by the average of the sum of the individual volumes.
$\zeta(t)$ is an epoch dependent weighting function that follows a pre-determined schedule.
At the beginning of training, we want the network to focus on alignment so $\zeta(t)$ was initialized with a small value. 
However, at later epochs, $\zeta(t)$ was gradually increased, so toward the end of training reconstruction quality was emphasized more in order to obtain a good shape descriptor. 
The first term in Eq.~\ref{eq:loss} is the alignment loss, whereas the second term is the reconstruction loss.

Although the user can pick an arbitrary template $x_{ref}$ (such as some training sample), in our implementation we optimized it via minimizing the loss function. 
The learned template volume was passed through a sigmoid layer in order to ensure that the voxel values lie between zero and one. 
Similar to the widely used batch-norm layer, we introduced a parameter normalizer layer at the output of the STN that ensures that the (mini-batch) average value of each of the nine parameters (3 rotation angles, 3 translations and 3 log scales) is equal to zero.
This way, the learned template does not experience drift in rotation, scaling and translation over the training epochs.
Without a parameter normalizer, there's nothing in the learning dynamics that would prevent the learned template to continuously rotate, for example.

\section{Experiments and Results}
\label{sec:majhead}
\begin{figure}
    \centering
    \includegraphics[width=4cm]{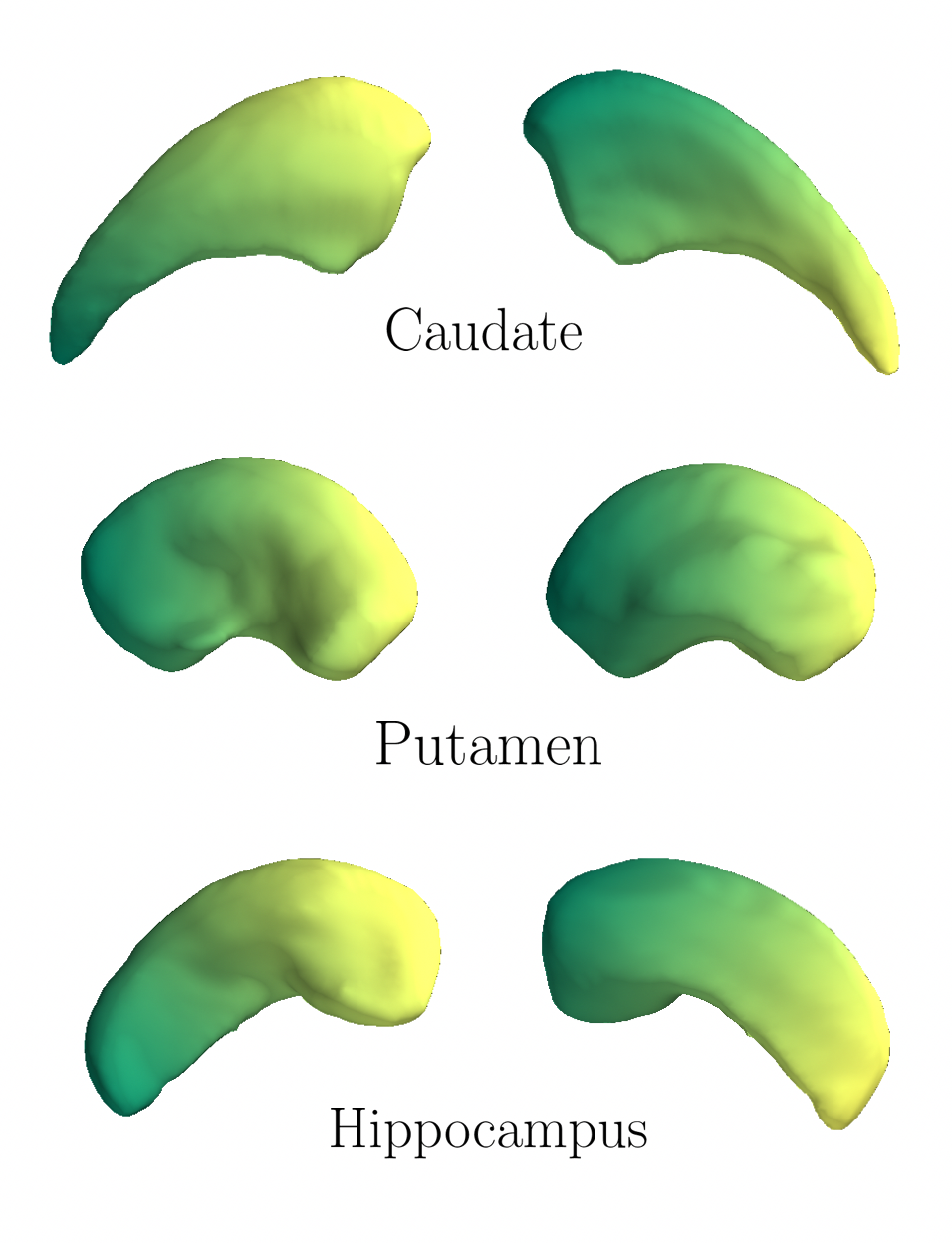}
    \caption{Lateral and medial views of the learned templates}
    \label{fig:learned_template}
\end{figure}

\subsection{Dataset}
\label{ssec:dataset}

To showcase and validate the proposed algorithm, we used healthy subjects from the OASIS-1 dataset \cite{oasis} \url{https://www.oasis-brains.org/}, spanning the ages of 18 through 96. 
The total sample size was 315. 
We split the data into three non-overlapping groups containing 165, 50, and 100 subjects.
These were used for training, validation, and testing, respectively. 
For each MRI scan, we extracted 6 brain regions: caudate, putamen and hippocampus in the two hemispheres. 
These structures were automatically segmented using FreeSurfer (v 5.1), which were visually inspected for quality assurance.
The left hemisphere ROIs were mirrored and combined with the right hemishpere data. 

There were 20 healthy subjects who obtained a second (repeat) scan on a subsequent visit within 90 days of their initial session.
These subjects were all included in our test dataset and the repeat scans were used for our retrieval experiment, as described below.

\begin{table}[]
\centering 
\scalebox{0.8}{
\begin{tabular}{c|c|c|c|c|}
\cline{2-5}
\multicolumn{1}{l|}{}                                                                                & \multicolumn{2}{c|}{Type}           & \multicolumn{1}{l|}{ShapeDNA} & Ours            \\ \hline
\multicolumn{1}{|c|}{\multirow{4}{*}{\rotatebox[origin=c]{90}{\begin{tabular}[c]{@{}c@{}}Same \\ Scan \end{tabular}}}}          & \multirow{2}{*}{Similarity} & Top 1 & \textbf{94.17}                & 91.58           \\ \cline{3-3}
\multicolumn{1}{|c|}{}                                                                               &                             & Top 5 & 97.17                         & \textbf{98.66}  \\ \cline{2-5} 
\multicolumn{1}{|c|}{}                                                                               & \multirow{2}{*}{Affine}     & Top 1 & 56.75                         & \textbf{93.67}  \\ \cline{3-3}
\multicolumn{1}{|c|}{}                                                                               &                             & Top 5 & 70.92                         & \textbf{98.61}  \\ \hline
\multicolumn{1}{|c|}{\multirow{6}{*}{\rotatebox[origin=c]{90}{\begin{tabular}[c]{@{}c@{}}Repeated \\ Subjects Only \end{tabular}}}} & \multirow{2}{*}{No}         & Top 1 & 55.83                         & \textbf{99.17}  \\ \cline{3-3}
\multicolumn{1}{|c|}{}                                                                               &                             & Top 5 & 81.67                         & \textbf{100} \\ \cline{2-5} 
\multicolumn{1}{|c|}{}                                                                               & \multirow{2}{*}{Similarity} & Top 1 & 40.00                         & \textbf{73.06}  \\ \cline{3-3}
\multicolumn{1}{|c|}{}                                                                               &                             & Top 5 & 70.83                         & \textbf{90.83}  \\ \cline{2-5} 
\multicolumn{1}{|c|}{}                                                                               & \multirow{2}{*}{Affine}     & Top 1 & 30.83                         & \textbf{78.33}  \\ \cline{3-3}
\multicolumn{1}{|c|}{}                                                                               &                             & Top 5 & 60.00                         & \textbf{93.89}  \\ \hline
\multicolumn{1}{|c|}{\multirow{6}{*}{\rotatebox[origin=c]{90}{\begin{tabular}[c]{@{}c@{}}Expanded \\ Look-Up \end{tabular}}}}   & \multirow{2}{*}{No}         & Top 1 & 42.50                         & \textbf{99.17}  \\ \cline{3-3}
\multicolumn{1}{|c|}{}                                                                               &                             & Top 5 & 65.00                         & \textbf{100} \\ \cline{2-5} 
\multicolumn{1}{|c|}{}                                                                               & \multirow{2}{*}{Similarity} & Top 1 & 28.33                         & \textbf{68.89}  \\ \cline{3-3}
\multicolumn{1}{|c|}{}                                                                               &                             & Top 5 & 55.83                         & \textbf{86.11}  \\ \cline{2-5} 
\multicolumn{1}{|c|}{}                                                                               & \multirow{2}{*}{Affine}     & Top 1 & 20.00                         & \textbf{72.78}  \\ \cline{3-3}
\multicolumn{1}{|c|}{}                                                                               &                             & Top 5 & 40.83                         & \textbf{90.83}  \\ \hline
\end{tabular}}
\caption{Results}
\label{table}
\end{table}

\subsection{Implementation Details of Proposed Approach}
We augmented our training data by randomly rotating up to 35$^{\circ}$ around all three axes. 
Similarly, we applied an axis-independent random scale up to $\pm 50\%$. 
We trained a single model for the three different structure types we considered in our experiments: caudate, putamen, and hippocampus.
However, we used a separate template for each structure, thus learning three templates.
The templates were initialized using a single, average 43-year old training subject. 
The learned templates are shown in Figure \ref{fig:learned_template}. 

We optimized our loss function using ADAM~\cite{kingma2014adam}, with stochastic gradients computed on a mini batch size of 12 (4 examples of each structure). 
$\zeta(t)$ was set to $10^{-10}$ during the first epoch, and increased up to $10^{-3}$ linearly with each epoch. 
The implementation is in PyTorch and the code is freely available at \url{https://github.com/evanmy/voxel_shape_analysis}.

\subsection{Benchmark Method}
We compared our method to ShapeDNA \cite{shapedna}, which uses a surface-based strategy to derive shape descriptors that are invariant to isometric transformations. 
ShapeDNA uses the Laplace-Beltrami spectrum of the surface mesh and has been successfully applied to the study of brain structures~\cite{brainprint}. 
`
\subsection{Retrieval Experiments}
A good shape descriptor should not only be invariant to specific transformations, but also capture meaningful differences across subjects while remaining stable for a given subject. 
To this end, we performed two retrieval experiments. 

In first experiment, we applied random transformations to test subjects (up to 15$^{\circ}$ rotation and 20\% scale on each axis, respectively) to create query images.
We considered two scenarios for the transformations: Similarity and Affine.
In the Similarity case, we applied random rotations coupled with global scales.
In the Affine case, each axis was randomly scaled independently, in addition to the random rotations.

The L2-norm was computed between the representation derived for the randomly transformed query image and the representations from all original images of the test subjects. 
We then ranked these distances and report the top-1 and top-5 accuracy values in Table \ref{table}. 
Top-X refers to the fraction of query instances where the query subject ranked among the top X smallest L2-distances in shape space. 


In the second experiment, we used the repeat scans from the 20 subjects in the test set. 
Similar to above, the shape descriptor derived from the repeat scans should lie close to the first scans of the corresponding subjects. 
We considered three transformation scenarios. 
No transformation, random similarity and random affine, where the random transformations were implemented as in the first experiment.
We also considered two retrieval scenarios. 
In first case (Repeat Subjects Only), the look-up dataset consisted only of the 20 test subjects with repeat scans.
In the second case (Expanded Look-Up), the look-up dataset included all 100 test subjects.
We computed top-1 and top-5 accuracy values for these different scenarios, reported in Table \ref{table}.

In first experiment, ShapeDNA yielded high accuracy for similarity transformations, which is unsurprising since it is isometry invariant. However, ShapeDNA performed poorly with affine transformations, whereas the proposed method achieved high accuracy for both types of transformations. 
In the second experiment with the repeat scans, our algorithm vastly outperformed the benchmark, under all considered scenarios. 
Since our method works directly in voxel space, it is not as sensitive to the mesh surface that is required for the ShapeDNA benchmark.
We believe that this makes our method more stable across repeat scans of the same subject.

\section{Conclusion}
We proposed a data driven method to learn a 3D shape descriptor. 
Geometric transformations are normalized for through the spatial transformer by aligning the input to a learned template. 
Furthermore, a concise shape descriptor is obtained through an autoencoder.  
Our method outperforms an existing benchmark on  retrieval experiments of subjects with longitudinal scans. 
Future work will include visualizing the learned shape descriptor and its application to examining associations between genetic/clinical variables and neuroanatomical shape.



\bibliographystyle{IEEEbib}
\bibliography{strings,refs}

\end{document}